\documentclass[runningheads]{llncs}

\usepackage{eccv}

\usepackage{eccvabbrv}

\usepackage{graphicx}
\usepackage{booktabs}

\usepackage[accsupp]{axessibility}  % Improves PDF readability for those with disabilities.

\usepackage{multirow}
\usepackage{booktabs}
\usepackage{colortbl}

\usepackage[pagebackref,breaklinks,colorlinks,citecolor=eccvblue]{hyperref}
\usepackage{orcidlink}

\begin{document}

\title{Bridging Different Language Models and \\ Generative Vision Models for \\ Text-to-Image Generation} 

\titlerunning{LaVi-Bridge}

\author{Shihao Zhao\inst{1} \and
Shaozhe Hao\inst{1} \and
Bojia Zi\inst{2} \and
Huaizhe Xu\inst{3} \and
Kwan-Yee K. Wong\inst{1}
}

\authorrunning{S. Zhao et al.}

\institute{The University of Hong Kong \\
\email{\{shzhao,szhao,kykwong\}@cs.hku.hk} \and
The Chinese University of Hong Kong\\
\email{bjzi@se.cuhk.edu.hk} \and
The Hong Kong University of Science and Technology \\
\email{hxubr@connect.ust.hk}}

\maketitle

\begin{abstract}

Text-to-image generation has made significant advancements with the introduction of text-to-image diffusion models. These models typically consist of a language model that interprets user prompts and a vision model that generates corresponding images. As language and vision models continue to progress in their respective domains, there is a great potential in exploring the replacement of components in text-to-image diffusion models with more advanced counterparts. A broader research objective would therefore be to investigate the integration of any two unrelated language and generative vision models for text-to-image generation. In this paper, we explore this objective and propose LaVi-Bridge, a pipeline that enables the integration of diverse pre-trained language models and generative vision models for text-to-image generation. By leveraging LoRA and adapters, LaVi-Bridge offers a flexible and plug-and-play approach without requiring modifications to the original weights of the language and vision models. Our pipeline is compatible with various language models and generative vision models, accommodating different structures. Within this framework, we demonstrate that incorporating superior modules, such as more advanced language models or generative vision models, results in notable improvements in capabilities like text alignment or image quality. Extensive evaluations have been conducted to verify the effectiveness of LaVi-Bridge. Code is available at \url{https://github.com/ShihaoZhaoZSH/LaVi-Bridge}.


\end{abstract}

\section{Introduction}
\label{section:Introduction}

\begin{figure}[t]
\centering
  \includegraphics[width=1\linewidth]{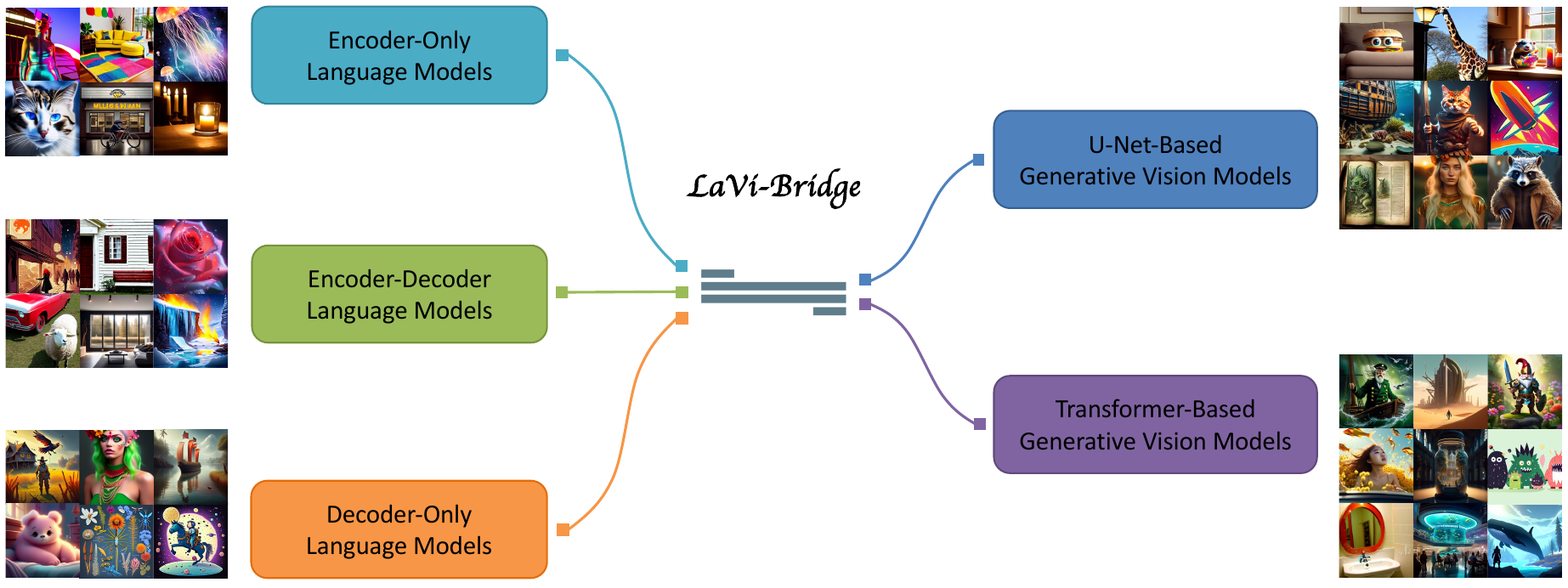}
  \caption{Overview of LaVi-Bridge. LaVi-Bridge is capable of integrating various language models and generative vision models. On the left side, we keep the vision model fixed and experiment with different language models in our pipeline. On the right side, we keep the language model fixed and try out different vision models. We display the visualization results alongside each combination.}
  \label{figure: fig1}
\end{figure}

In recent years, there have been remarkable advancements in the field of text-to-image generation, specifically through the use of diffusion models \cite{ho2020denoising,song2020denoising,song2020score,dhariwal2021diffusion}. These models have made significant contributions and have gained considerable attention for their exceptional performance. By leveraging large-scale training datasets alongside large deep models, text-to-image diffusion models are capable of producing high-quality images that faithfully align with the textual descriptions provided by users. This has rendered them highly applicable in real-world scenarios such as content creation and architectural design.

Text-to-image diffusion models \cite{nichol2021glide,rombach2022high,saharia2022photorealistic,ramesh2022hierarchical,Dalle-3,xue2024raphael,chen2023pixart} typically consist of two key components, namely a language model and a generative vision model. The language model is responsible for comprehending the input prompts, whereas the vision model is tasked with generating images that align with the extracted context. Existing text-to-image diffusion models employ various language models and generative vision models and have gained widespread usage. For instance, Stable Diffusion (SD) \cite{rombach2022high} is a highly popular text-to-image diffusion model that employs the CLIP text encoder \cite{radford2021learning} as its language model and a U-Net \cite{ronneberger2015u} as its generative vision model. Another example is PixArt \cite{chen2023pixart}, a recently proposed text-to-image diffusion model that adopts the T5 \cite{raffel2020exploring} as its language model and a Vision Transformer (ViT) \cite{dosovitskiy2020vit} as its generative vision model. These models are trained on a vast amount of text-image pairs, enabling seamless collaboration between their language modules and vision modules.

The advancements in deep language models and deep vision models have witnessed rapid progress in recent years, with both fields experiencing continuous developments and the introduction of more powerful models. However, this rapid development poses a challenge for the research in text-to-image generation when it comes to integrating more advanced language or vision models into existing text-to-image diffusion models. The problem of how to integrate any two unrelated language and vision models is unexplored, and the impact of newly developed models on text-to-image generation capabilities also remains uncertain. The current situation highlights the presence of a gap between the language or vision modules within text-to-image diffusion models and the state-of-the-art models in their respective domains. Therefore, it has become crucial to address this gap and explore ways to incorporate more advanced language or vision models into existing text-to-image diffusion models. Furthermore, the broader challenge of integrating any pre-trained language model with any generative vision model deserves a thorough investigation.

In this paper, our objective is to delve into the aforementioned problem. We propose LaVi-Bridge, a flexible framework that facilitates the integration of diverse well-trained language models and generative vision models to achieve text-to-image generation. Our framework enables the integration of two unrelated language and vision models that have not been previously trained together, as shown in \cref{figure: fig1}. Importantly, LaVi-Bridge does not require modifying the original weights of the language and vision models. Instead, it injects LoRA \cite{hu2021lora} into the language and vision models separately and utilizes an adapter to bridge these two modules. Moreover, LaVi-Bridge only necessitates a relatively small dataset to integrate different language models and generative vision models for text-to-image generation.

We summarize the advantages and features of LaVi-Bridge as follows:
\begin{enumerate}
  \item LaVi-Bridge is designed for text-to-image diffusion models and serves as a bridge, capable of connecting various pre-trained language models and generative vision models. Our framework can accommodate different model structures, including encoder-only, encoder-decoder, and decoder-only language models, as well as U-Net-based and Transformer-based generative vision models.
  \item LaVi-Bridge utilizes LoRA and adapters, eliminating the need to modify the original weights of the models. It is more flexible and requires relatively small computing resources compared to training the entire diffusion model.
  \item We evaluated various text-image alignment and image quality metrics on short prompts, long prompts, and compositional prompts. We also conducted extensive visualization. We then drew several conclusions. For instance, integrating superior models leads to improved performance in the corresponding modality, such as enhanced semantic understanding with advanced language models or improved image quality with more powerful generative vision models. Additionally, the diffusion model utilizing Llama-2 demonstrates exceptional semantic understanding, while the diffusion model utilizing the transformer in PixArt yields images with enhanced aesthetics.
\end{enumerate}

\section{Related Work}
\label{section:Related Work}

\subsection{Language Models and Generative Vision Models}
\label{Subsection: Language Models and Generative Vision Models}

The mainstream Large Language Models (LLMs) \cite{devlin2018bert,raffel2020exploring,radford2018improving,radford2019language} are built based on the transformer structure \cite{vaswani2017attention}, with three main types of architectures, namely encoder-only, encoder-decoder, and decoder-only. All these three belong to Sequence to Sequence (Seq2Seq) \cite{sutskever2014sequence}. The encoder-only architecture is exemplified by BERT \cite{devlin2018bert}. CLIP text encoder \cite{radford2021learning} is based on BERT and further trained to align with the image domain. Models of this type excel at understanding the content of the input and generating outputs tailored to specific tasks. On the other hand, the encoder-decoder framework is adept at handling tasks that involve complex mappings between input and output sequences. Examples include T5 \cite{raffel2020exploring} and BART \cite{lewis2019bart}. Recently, due to the tremendous success of ChatGPT, attention has been drawn to models that consist solely of a decoder, like GPT-3 \cite{brown2020language} and Llama-2 \cite{touvron2023llama}. The decoder-only architecture demonstrates exceptional performance in semantic understanding. For text-to-image generation, all three types of LLMs can provide effective semantic information to serve as conditions for image generation in diffusion models. In this paper, we explore and compare all these three types of language models.

A generative vision model refers to a vision model with the ability to generate images or visual contents. There are two common types of structures, namely U-Net-based \cite{ronneberger2015u} and Transformer-based \cite{dosovitskiy2020vit}. Generative Adversarial Networks (GANs) \cite{goodfellow2014generative,reed2016generative,zhang2017stackgan} employ a framework consisting of a discriminator and a generator, with the generator's structure based on U-Net. On the other hand, motivated by the success of GPT models, recent works have attempted to use the Transformer architecture for image generation in an autoregressive manner, with notable examples being DALLE \cite{ramesh2021zero} and CogView \cite{ding2021cogview}. Another popular class of generative models is diffusion models \cite{ho2020denoising,song2020denoising,croitoru2023diffusion,song2020score}, which are based on the diffusion process and gradually denoise to produce natural images. Early diffusion models often employed U-Net as their generative vision model, such as Stable Diffusion, which scaled up the Latent Diffusion Model (LDM) ~\cite{rombach2022high} with larger data scales. Some recent works have started to replace the U-Net in diffusion models with Vision Transformer and have made significant progress, such as DiT \cite{peebles2023scalable}, U-ViT \cite{bao2022all} and PixArt \cite{chen2023pixart}. In this paper, we focus on diffusion models and explore both U-Net-based and Transformer-based vision models.

\subsection{Text-to-Image Diffusion Models}
\label{Subsection: Text-to-Image Diffusion Models}

Text-to-image diffusion models \cite{dhariwal2021diffusion,nichol2021glide,rombach2022high,saharia2022photorealistic,Dalle-3,song2023consistency} are capable of generating images based on user prompts. These models consist of two main components, namely a language model and a vision model. The language module is responsible for understanding the text input provided by the user, extracting contextual information, and injecting it into the vision module to generate the desired image. Text-to-image diffusion models have paved the way for various exciting research areas, including image editing \cite{meng2021sdedit,brooks2023instructpix2pix,kawar2023imagic}, controllable image generation \cite{zhang2023adding,zhao2024uni,mou2023t2i}, personalized object generation \cite{gal2022image,ruiz2023dreambooth,hao2023vico}, as well as other interesting applications \cite{guo2023animatediff,ge2023expressive,chen2024textdiffuser}. In the extensive exploration of diffusion models, researchers have utilized different language models and vision models. For instance, Stable Diffusion \cite{rombach2022high} employs CLIP text encoder \cite{radford2021learning} as its language model and U-Net as its vision model. Imagen \cite{saharia2022photorealistic} utilizes T5 \cite{raffel2020exploring} as its language model, which claims to enhance both sample fidelity and image-text alignment. ParaDiffusion \cite{wu2023paragraph} focuses on paragraph-to-image generation and leverages the powerful semantic understanding capability of Llama-2 \cite{touvron2023llama} to comprehend lengthy sentences. PixArt \cite{chen2023pixart}, on the other hand, utilizes a ViT \cite{dosovitskiy2020vit} as its vision model and achieves high image fidelity while being trained at a lower cost.

After training on a large dataset of text-image pairs \cite{schuhmann2021laion}, the language and vision models in the text-to-image diffusion model become closely intertwined. This tight coupling ensures a strong alignment between the provided text description and the generated image, but at the same time also limits the flexibility of the diffusion model. For instance, if a more advanced language or vision model becomes available, it may have the potential to enhance the text-to-image task. However, decoupling the language and vision modules in existing text-to-image diffusion models and replacing a module with a new one is nontrivial. Therefore, this paper explores the dilemma faced by text-to-image generation and proposes a framework that enables efficient integration of various language models and generative vision models.

\section{Method}
\label{section:Method}

\subsection{Preliminary}
\label{Subsection:Preliminary}

A diffusion model is based on the diffusion process for image generation. This process consists of two stages, namely the forward process and the reverse process. During the forward process, Gaussian noise is progressively added to a natural image until the image becomes completely noisy. After that, during the reverse process, the noise is gradually eliminated over a series of time steps, resulting in a natural image. In the reverse process, a trainable vision model is used to predict and remove the noise. By employing this denoising model, we are able to obtain a natural image from Gaussian noise by denoising. Within a text-to-image diffusion model, there are two components at each time step, namely a language model $f$ and a vision model $g$. The language model converts user input text $y$ into embeddings, which capture the semantic meaning of the text. On the other hand, the vision model, which is the denoising model aforementioned, encodes image features $z$, extracting relevant visual information from the input images. The interaction between text embeddings and image features is achieved through cross-attention layers, which can be formulated as
\begin{align}
\label{equation: 1}
    c=f(y), \\[6pt]
    Q=W_q(z), K=W_k&(c), V=W_v(c), \\[6pt]
    CrossAttention(Q, K, V)&=softmax(Q\cdot K^{T})\cdot V, 
\end{align}
where $W_q, W_k$ and $W_v$ are projection matrices.

\subsection{Language and Vision Alignment}
\label{Subsection: Language and Vision Alignment}

LaVi-Bridge enables the integration of any two pre-trained language and generative vision models, even though these models are not related and have been trained separately. Here, we denote the language model as $f$ and the vision model as $g$, as mentioned previously. If we directly interact the textual information and image information using \cref{equation: 1}, considering that $f$ and $g$ are trained independently, the parameters in the cross-attention layers of $g$ cannot comprehend the text embedding output by $f$, resulting in meaningless model outputs. 

To establish a connection between them, LaVi-Bridge keeps the pre-trained language and vision models fixed and utilizes LoRA to introduce trainable parameters $\Delta \theta$ into both the language model and the vision model. In this context, we denote the language and vision models with LoRA as $f^{\theta_{1} + \Delta \theta_{1}}$ and $g^{\theta_{2} + \Delta \theta_{2}}$, where $\theta_{1}$ and $\theta_{2}$ are the original parameters of $f$ and $g$, respectively. Furthermore, we introduce an adapter as a bridge between the language model and vision model to facilitate better alignment. The adapter consists of stacked feedforward layers, denoted as $h$. Consequently, the cross-attention layer can be expressed as 
\begin{align}
\label{equation: 2}
    c=&f^{\theta_{1} + \Delta \theta_{1}}(y), \\[6pt]
    Q=W_{q}^{\theta_{2} + \Delta \theta_{2}}(z), K=W_{k}&^{\theta_{2} + \Delta \theta_{2}}(h(c)), V=W_{v}^{\theta_{2} + \Delta \theta_{2}}(h(c)), \\[6pt]
    CrossAttention(Q, K, &V)=softmax(Q\cdot K^{T})\cdot V.
\end{align}

\begin{figure}[t]
\centering
  \includegraphics[width=1\linewidth]{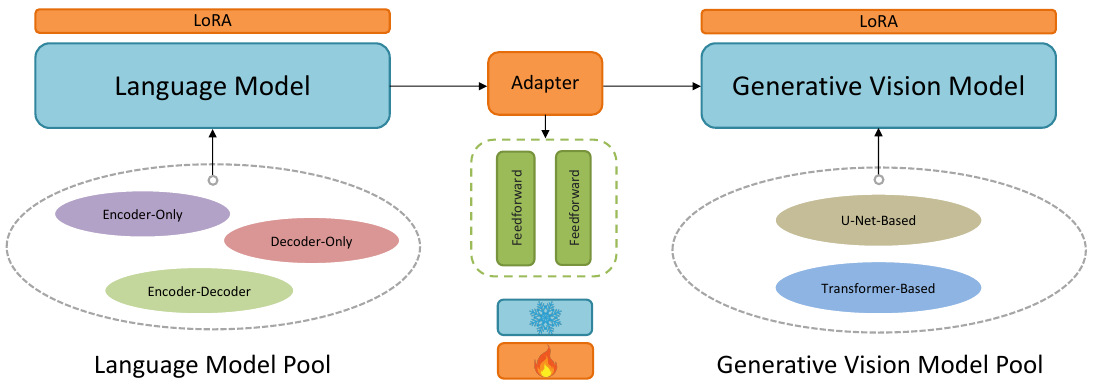}
  \caption{Pipeline of LaVi-Bridge. We select one model each from the language and vision model pools. We then freeze the pre-trained language and vision models and incorporate LoRA into both models. The connection between the language and vision models is established through an adapter. The only weights we need to train are the ones introduced by LoRA and the adapter.}
  \label{figure: fig2}
\end{figure}

Now, we only need to train $\Delta \theta_{1}$, $\Delta \theta_{2}$, and $h$ on a relatively small amount of text-image pairs. After training, the language and generative vision models can effectively collaborate to generate meaningful images. We present the framework of LaVi-Bridge in \cref{figure: fig2}. LaVi-Bridge is very straightforward, with both LoRA and the adapter being its crucial and indispensable components. 

\subsection{Design Details}
\label{Subsection: Detailed Design}

LaVi-Bridge is designed to accommodate a wide range of language model structures, including encoder-only, encoder-decoder, decoder-only, as well as generative vision model structures such as U-Net and ViT. In the language model, we inject LoRA into all linear layers of the attention layers. Likewise, in a transformer-based vision model, LoRA is injected into all linear layers of the attention layers. In a U-Net-based vision model, LoRA is injected into all linear layers and convolutional layers of the ResBlocks, attention, and cross-attention layers. To address the dimension disparity between the output embedding of the language model and the dimensions handled by the cross-attention of the vision model, we employ two feedforward layers for the adapter. The input dimension of the adapter matches the output text embedding dimension of the language model, while the output dimension aligns with the dimensions received by the cross-attention of the vision model. 

For training, we first select the language and generative vision models that we choose to integrate. We keep their original weights fixed and train LoRA and the adapter on text-image pairs following the design mentioned above. The trained LoRA and adapter have fewer parameters compared to the original model weights, which makes LaVi-Bridge highly flexible. For evaluation, we used various metrics to assess text alignment and image quality across short prompts, long prompts, and compositional prompts.

\section{Experiments}
\label{section: Experiments}

\subsection{Experimental Settings}
\label{Subsection: Experimental Settings}
In this section, we explored the performance of different language models and generative vision models under LaVi-Bridge. We also tested the impact of LoRA and adapters. We trained on a dataset consisting of a total of 1 million text-image pairs, including around 600k text-image pairs from the COCO2017 \cite{lin2014microsoft} train set and 400k text-image pairs from an internal dataset with high-quality images and captions. For each setting, we set the LoRA rank to $32$, image resolution to $512 \times 512$ and the batch size to $256$. We used the AdamW optimizer \cite{loshchilov2017decoupled} with a learning rate of $1 \times 10^{-4}$ and trained for a total of $50k$ steps. During inference, we employed the DDIM sampler \cite{song2020denoising} for sampling with the number of time steps set to $50$ and the classifier free guidance scale \cite{ho2021classifierfree} set to $7.5$.

As mentioned above, we conducted our quantitative evaluation on short prompts, long prompts, and compositional prompts. Specifically, 
\begin{enumerate}
  \item For short prompts, we evaluated using the COCO2014 \cite{lin2014microsoft} validation set. We randomly sampled 30k images and tested image quality and text alignment within this subset. We used FID \cite{heusel2017gans} and aesthetic score \cite{aesthetic-predictor} as evaluation metrics for image quality and CLIP score for text alignment.
  \item For long prompts, we employed the same 30k-subset of COCO2014 and utilized Llama-2 to generate expanded captions ranging from $20$ to $70$ words to construct a dataset of 30k long prompts. Since the caption expansion process does not refer to the content of the reference image, we solely used aesthetic score to evaluate image quality and CLIP score for text alignment.
  \item For compositional prompts, we utilized the benchmark proposed by Compbench \cite{huang2023t2icompbench}. Compositional prompts were mainly used to test the model's understanding of textual attributes, such as generating correct object properties like color and shape, as well as accurate relationships between objects, such as spatial positioning. 
\end{enumerate}
We conducted a user study on different combinations of language and vision models. For each combination, we evaluated two metrics, namely image quality and text alignment. Users were asked to rank the generated images based on these evaluation criteria. The image ranked last received a score of $1$, the second-to-last received a score of $2$, and so on. We then calculated the percentage of scores for each model. We selected $20$ prompts and included $30$ users participated in the testing. In addition to quantitative evaluation and user study, we provided ample visualization results in each section to offer a more intuitive understanding of the performance of each model.

\begin{figure}[t]
\centering
  \includegraphics[width=1\linewidth]{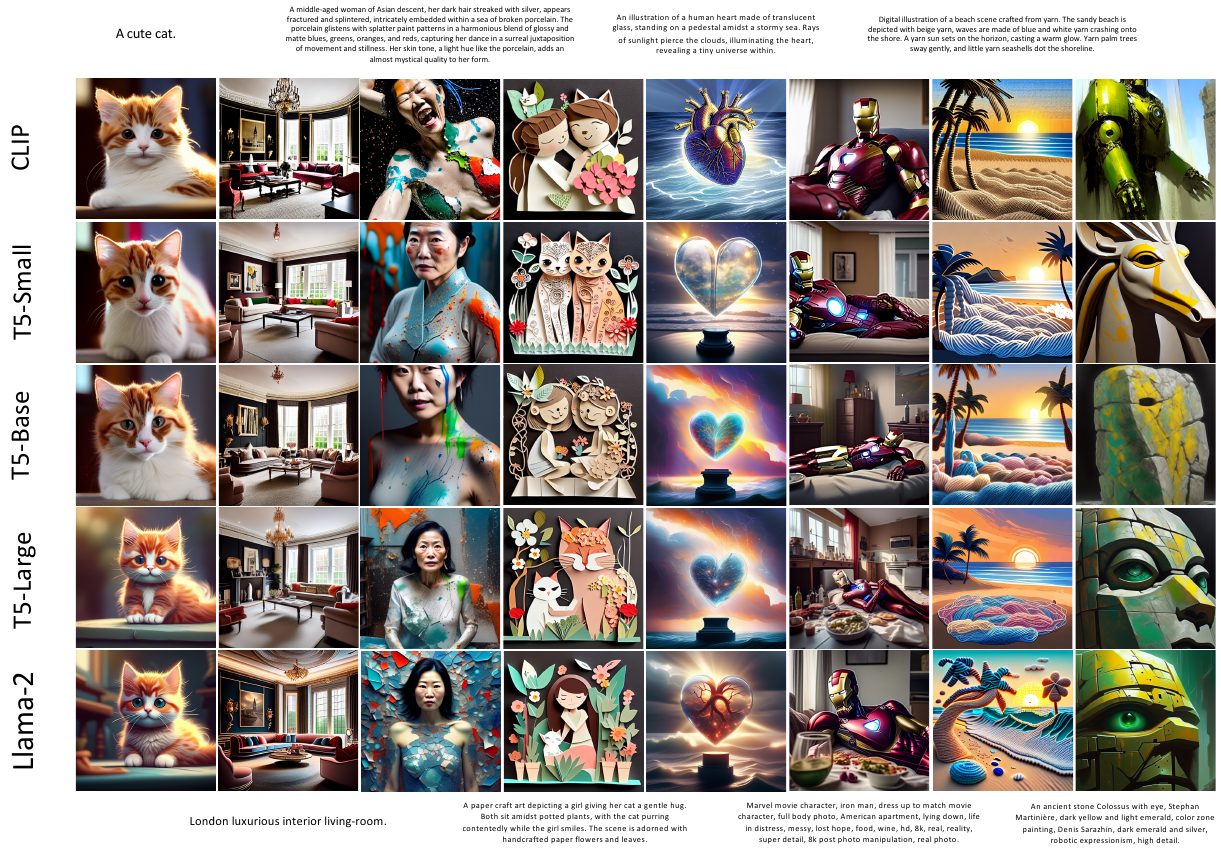}
  \caption{Visualization results of LaVi-Bridge with different language models. The first row to the fifth row present the results with CLIP text encoder, T5-Small, T5-Base, T5-Large, and Llama-2, respectively. The prompts are displayed at the top or bottom of each column.}
  \label{figure: fig3}
\end{figure}

\subsection{Evaluation on Different Language Models}
\label{Subsection: Evaluation on Different Language Models}

\begin{table}[t]
\small
\centering
  \setlength\tabcolsep{7.5pt}
  \caption{Quantitative evaluation of LaVi-Bridge with different language models. ``Short'', ``Long'' and ``Comp'' denote short prompts,  long prompts, and compositional prompts respectively. The best results are in \textbf{bold}.}
  \label{table: 1}
  \centering
  \scalebox{1}{
  \begin{tabular}{lccccc}
    \toprule
    & CLIP & T5-Small & T5-Base & T5-Large & Llama-2 \\
    \midrule
    Short - FID  & 23.57 & 22.98 & 22.62 & 23.11 & \textbf{21.80} \\
    Short - Aesthetics & 5.609 & 5.813 & \textbf{5.888} & 5.881 & 5.883 \\
    Short - CLIP Score & 0.3102 & 0.3122 & 0.3149 & 0.3156 & \textbf{0.3172} \\
    \midrule
    Long - Aesthetics & 6.003 & 6.206 & 6.284 & 6.305 & \textbf{6.355} \\
    Long - CLIP Score & 0.3120 & 0.3111 & 0.3179 & 0.3193 & \textbf{0.3231} \\
    \midrule
    Comp - Color & 0.3578 & 0.3368 & 0.3856 & 0.3889 & \textbf{0.4859} \\
    Comp - Shape & 0.3752 & 0.2962 & 0.3266 & 0.3552 & \textbf{0.4285} \\
    Comp - Texture & 0.4506 & 0.3728 & 0.4132 & 0.4524 & \textbf{0.5055} \\
    Comp - Spatial & 0.1296 & 0.1456 & 0.1569 & 0.1582 & \textbf{0.1914} \\
    Comp - Non-Spatial & 0.3009 & 0.2984 & 0.3054 & 0.3068 & \textbf{0.3106} \\
    Comp - Complex & 0.2985 & 0.2728 & 0.3055 & 0.3072 & \textbf{0.3094} \\
    \bottomrule
  \end{tabular}}
\end{table}

This section evaluates the performance of LaVi-Bridge with different language models. We fixed the vision model to the U-Net of Stable Diffusion V1.4 and integrated it with different language models under LaVi-Bridge. We considered CLIP text encoder, based on the encoder-only framework, T5 series (T5-Small, T5-Base, T5-Large), based on the encoder-decoder framework, and Llama-2-7B, based on the decoder-only framework. We present the visualization results in \cref{figure: fig3}, quantitative evaluation in \cref{table: 1}, and user study in \cref{figure: fig5}.

\noindent \textbf{Visualization} 
From \cref{figure: fig3}, we can observe that with LaVi-Bridge, all these language models can effectively integrate with U-Net of Stable Diffusion V1.4 and generate meaningful results, such as cases of the cat and living room in \cref{figure: fig3}. This demonstrates the great generalization ability of LaVi-Bridge for various language models. Additionally, we notice that the performance of different model structures varies when the provided prompts contain more complex semantics. We find that the text-to-image diffusion model corresponding to Llama-2 can perfectly describe semantic information. For example, in the third column, Llama-2's generated result effectively integrates a woman into the sea of fragmented porcelain. In the fourth column, it correctly understands and generates both the girl and the cat in a paper craft art. In the seventh column, it even portrays an entire beach scene using yarn. These examples surpass the capabilities of those models with CLIP and T5. Furthermore, we observe that T5-Large and Llama-2 accurately generate food and wine in the case of Iron Man, and in the last column, they successfully generate ``an ancient stone with eyes in dark yellow and emerald''. Models with CLIP text encoder, T5-Small, and T5-Base are not able to capture these cases accurately.

\noindent \textbf{Quantitative Evaluation} 
From \cref{table: 1}, we can observe that Llama-2 achieves the best results for all the metrics used to evaluate text alignment ability, under the setting of all the short prompts, long prompts, and compositional prompts. Besides, Llama-2 also performs the best on most of the metrics evaluating image quality. On the other hand, as the model capacity increases, in general circumstances T5-Large usually outperforms T5-Base, and T5-Base outperforms T5-Small in the area of Natural Language Processing. This conclusion also holds true for LaVi-Bridge. For all the metrics used to evaluate text alignment ability in \cref{table: 1}, T5-Large is superior to T5-Base, and T5-Base is superior to T5-Small. This tells us that incorporating a better language model into the text-to-image diffusion model under LaVi-Bridge can lead to improved text alignment. This makes one of the motivations of LaVi-Bridge meaningful, which is that replacing the model in the existing text-to-image diffusion model with a better model can lead to performance improvements.

\noindent \textbf{User Study} 
We follow the settings described in \cref{Subsection: Experimental Settings}, and the results are presented in the two disk diagrams on the left side of \cref{figure: fig5}. The model using Llama-2 demonstrates the best performance in terms of both image quality and text alignment, with a particularly pronounced advantage in text alignment. On the other hand, CLIP and T5-Small exhibit noticeably poorer performance on both image quality and text alignment compared to other models.

\subsection{Evaluation on Different Vision Models}
\label{Subsection: Evaluation on Different Vision Models}

\begin{figure}[t]
\centering
  \includegraphics[width=1\linewidth]{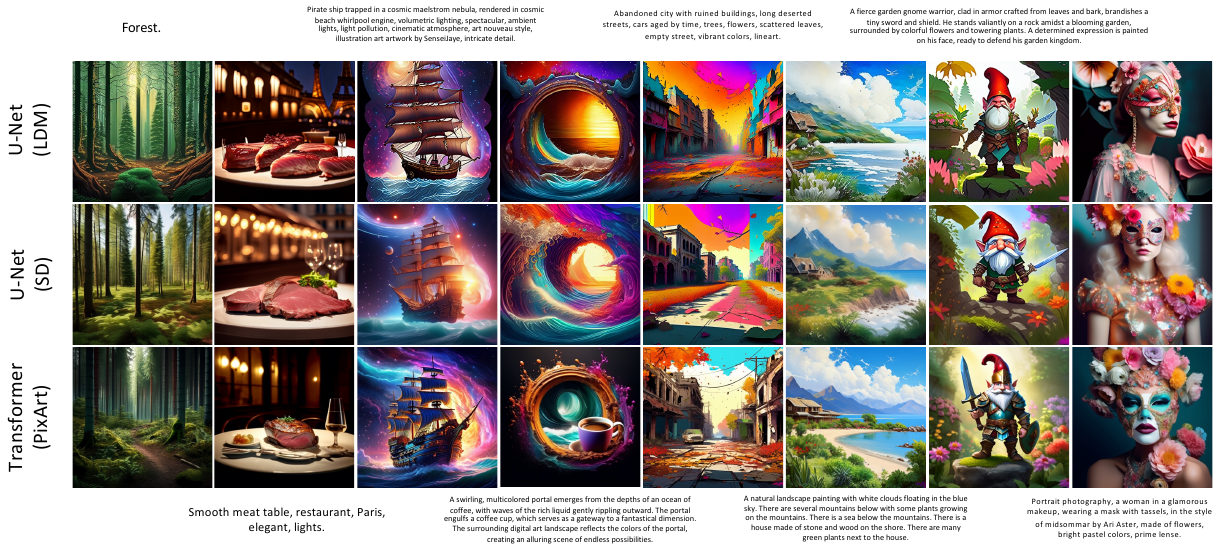}
  \caption{Visualization results of LaVi-Bridge under different generative vision models. The first row to the third row present the results with U-Net in Latent Diffusion Model, U-Net in Stable Diffusion V1.4 and transformer in PixArt, respectively. The prompts are displayed at the top or bottom of each column.}
  \label{figure: fig4}
\end{figure}

\begin{table}[t]
\small
\centering
  \setlength\tabcolsep{9pt}
  \caption{Quantitative evaluation of LaVi-Bridge under different generative vision models. ``Short'', ``Long'' and ``Comp'' denote short prompts,  long prompts, and compositional prompts respectively. The best results are in \textbf{bold}.}
  \label{table: 2}
  \centering
  \scalebox{1}{
  \begin{tabular}{lccc}
    \toprule
    & U-Net(LDM) & U-Net(SD) & Transformer(PixArt) \\
    \midrule
    Short - FID  & 25.94 & 23.11 & \textbf{23.02} \\
    Short - Aesthetics & 5.703 & 5.881 & \textbf{6.145} \\
    Short - CLIP Score & 0.3126 & 0.3156 & \textbf{0.3172} \\
    \midrule
    Long - Aesthetics & 6.122 & 6.305 & \textbf{6.406} \\
    Long - CLIP Score & 0.3189 & 0.3193 & \textbf{0.3210} \\
    \midrule
    Comp - Color & \textbf{0.4099} & 0.3889 & 0.3689 \\
    Comp - Shape & \textbf{0.3724} & 0.3552 & 0.3316 \\
    Comp - Texture & \textbf{0.5046} & 0.4524 & 0.4553 \\
    Comp - Spatial & 0.1550 & 0.1582 & \textbf{0.1725} \\
    Comp - Non-Spatial & 0.3004 & 0.3068 & \textbf{0.3098} \\
    Comp - Complex & 0.3060 & \textbf{0.3072} & 0.3014 \\
    \bottomrule
  \end{tabular}}
\end{table}

\begin{figure}[t]
\centering
  \includegraphics[width=1\linewidth]{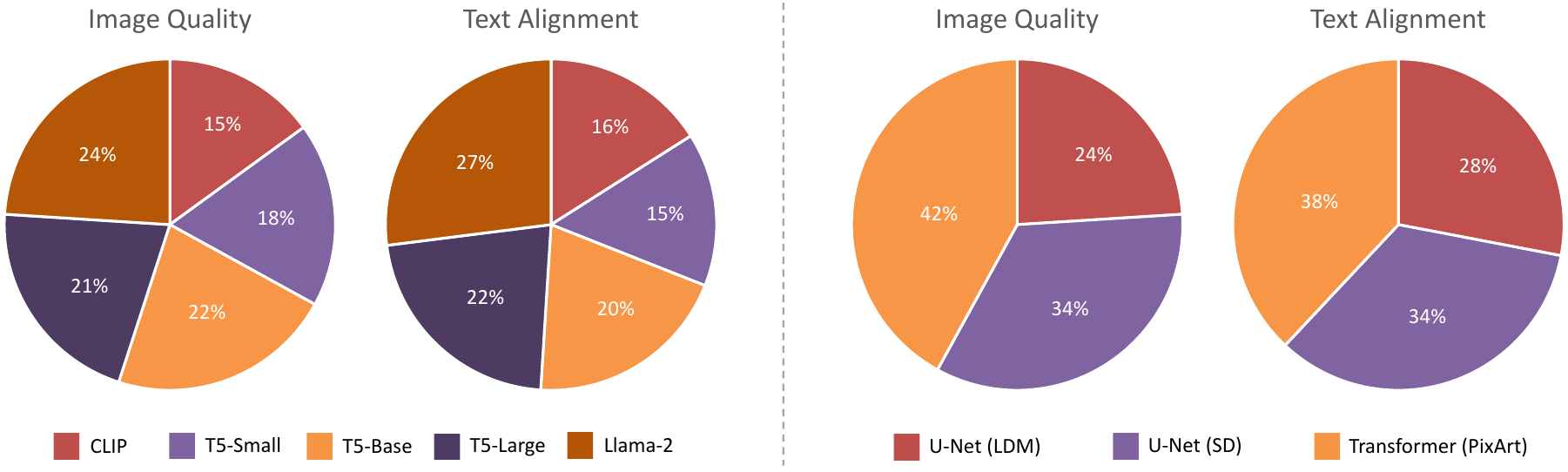}
  \caption{User study. The two disk diagrams on the left display the user's scoring results on different language models, while the two disk diagrams on the right display the user's scoring results on different generative vision models. The percentage represents the proportion of the score obtained by a model out of the total score of all models.}
  \label{figure: fig5}
\end{figure}

This section evaluates the performance of LaVi-Bridge with different vision models. We fixed the language model to T5-Large and integrated it with different generative vision models under LaVi-Bridge. We considered the well-trained U-Nets in the Latent Diffusion Model and Stable Diffusion V1.4, as well as the Vision Transformer in PixArt, totally three models. We present the visualization results in \cref{figure: fig4}, quantitative evaluation in \cref{table: 2}, and user study in \cref{figure: fig5}.

\noindent \textbf{Visualization}
From \cref{figure: fig4}, we can see that all these three vision models integrate well with T5-Large and generate relatively accurate images based on the given text prompts. From these cases, we can observe that the images generated by the transformer model based on PixArt exhibit richer details compared to the images generated by the other two models based on U-Net. For example, the forest in the first column, the hull of the pirate ship in the third column, and the bushes at the foot of the mountain in the sixth column are very intricate and realistic. Additionally, we can observe from these cases that the images generated by the model with U-Net of Stable Diffusion V1.4 have more detailed features compared to the images generated by the model with U-Net of Latent Diffusion Model. Furthermore, we can also find that, for the PixArt-based model, text alignment is better in some cases. For instance, in the image of the fifth column, only the model that is based on the transformer of PixArt generates the ``aged car'' mentioned in the prompt. Similarly, in the seventh column, the garden warrior holding a sword and shield is highly consistent with the prompt description. 

\noindent \textbf{Quantitative Evaluation}
From \cref{table: 2}, it can be observed that for all the metrics measuring image quality, LaVi-Bridge with the PixArt vision model achieves the best results. Additionally, PixArt also achieves the best text alignment for both short and long prompts. This reflects the use of PixArt's transformer as a vision model can also improves the model's understanding of semantics to some extent. Additionally, it is noteworthy that the U-Net in Stable Diffusion, an enhanced version of the U-Net utilized in the Latent Diffusion Model, still outperforms Latent Diffusion Model's U-Net under LaVi-Bridge on all the metrics measuring image quality. This aligns with our previous discussion in \cref{Subsection: Evaluation on Different Language Models} and further validates the underlying motivation behind our proposed LaVi-Bridge.

\noindent \textbf{User Study} 
We follow the settings described in \cref{Subsection: Experimental Settings}, and the results are presented in the two disk diagrams on the right side of \cref{figure: fig5}. The model using the transformer from PixArt demonstrates the best performance in terms of both image quality and text alignment, with a notably significant advantage in image quality. Additionally, the U-Net in Stable Diffusion outperforms the U-Net in the Latent Diffusion Model overall.

\subsection{Ablation Study}
\label{Subsection: Ablation Study}

\begin{figure}[t]
\centering
  \includegraphics[width=1\linewidth]{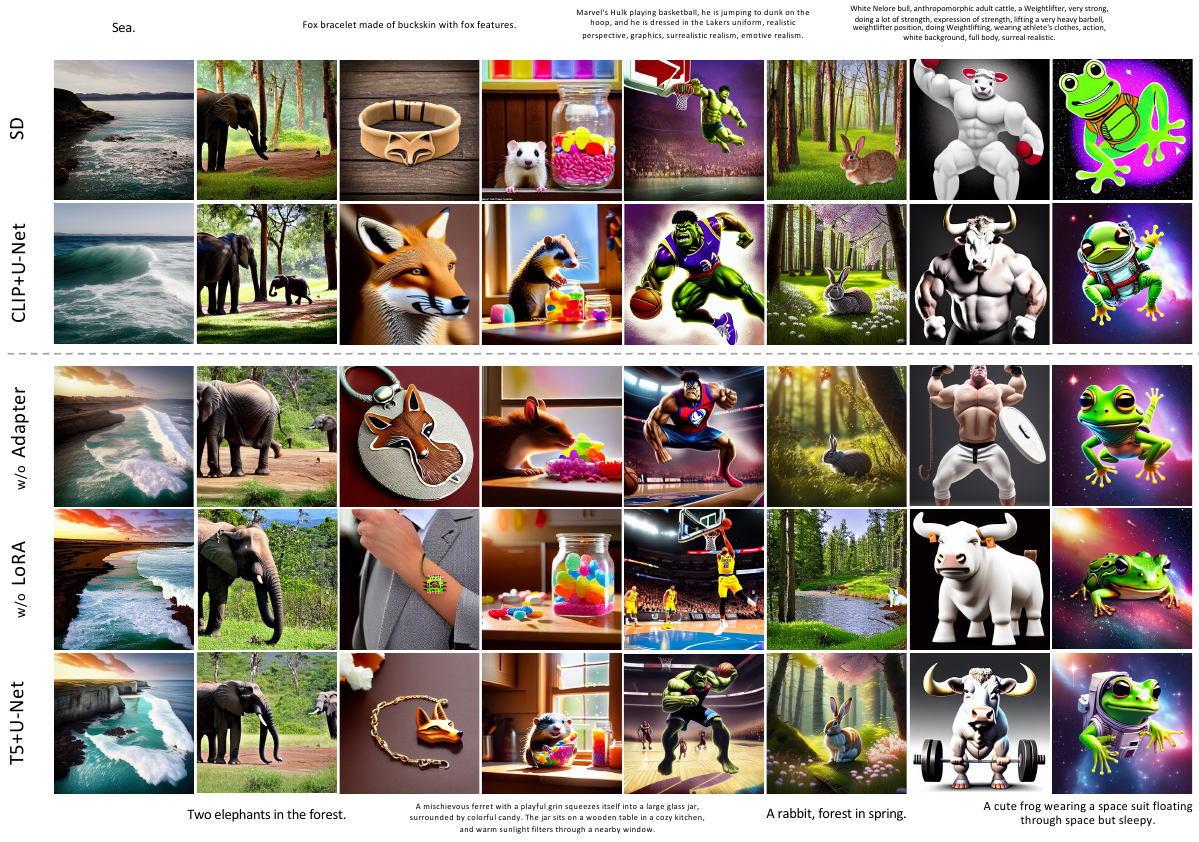}
  \caption{Visualization results of the ablation study. The top two rows show the impact of LaVi-Bridge on the original pre-trained text-to-image diffusion models. The bottom three rows illustrate the influence of the adapter and LoRA. The prompts are displayed at the top or bottom of each column.}
  \label{figure: fig6}
\end{figure}

In this section, we investigate two sets of ablation experiments. The first set aims to explore the impact of training LaVi-Bridge on the original pre-trained text-to-image diffusion model. The second set of experiments is to study the effects of LoRA and adapters in LaVi-Bridge. For both sets of experiments, we present visualization results in \cref{figure: fig6} and provide quantitative evaluations in \cref{table: 3}.

\noindent \textbf{Training with LaVi-Bridge}
We investigate the impact of our LaVi-Bridge training framework on the original pre-trained text-to-image diffusion model. Specifically, we consider Stable Diffusion V1.4 which adopts CLIP text encoder as its language model and U-Net as its vision model. We incorporate LoRA and an adapter and apply LaVi-Bridge to the same language and vision models with identical structures and weights to those in Stable Diffusion V1.4. We then compare the performance of the model under LaVi-Bridge with the original Stable Diffusion V1.4.

The visualization results are shown in the first two rows of \cref{figure: fig6}. For these two models, there is no significant difference in image quality and text alignment, varying on a case-by-case basis. In some cases, Stable Diffusion performs better, such as in the third column, where Stable Diffusion successfully generates the case of a ``Fox bracelet made of buckskin with fox features'', while the model trained under LaVi-Bridge only generates the fox and fails to understand the bracelet made of buckskin. Similarly, in the case of Marvel's Hulk playing basketball, Stable Diffusion generates a slam dunk action following the prompt, whereas the model trained under LaVi-Bridge does not. However, in the second column, the model trained under LaVi-Bridge correctly understands the quantity and successfully generates two elephants, while Stable Diffusion only generates one. Moreover, in the last column, the model trained under LaVi-Bridge accurately describes a frog in a spacesuit, while Stable Diffusion fails. 

The left two columns of \cref{table: 3} present the quantitative evaluation results. It can be observed that Stable Diffusion achieves the best image quality and text alignment for both short prompts and long prompts. However, for compositional prompts, the model trained under LaVi-Bridge outperforms Stable Diffusion in four out of six settings.

Based on the visualization results and quantitative evaluations, we can conclude that overall there is no significant improvement or decline in text alignment. Regarding image quality, it should be noted that training with LaVi-Bridge may result in a decrease compared to the original text-to-image diffusion model, if the same models and weights are used. However, it is important to understand that the main purpose of LaVi-Bridge is to establish connections between different language and vision models, enabling the utilization of more advanced models for performance enhancement. It is not intended to be directly applied to the original text-to-image diffusion models using the same models and weights.

\begin{table}[t]
\small
\centering
  \setlength\tabcolsep{2.5pt}
  \caption{Quantitative evaluation of the ablation study. The left two columns present the impact of LaVi-Bridge on the original pre-trained text-to-image diffusion models. The right three columns demonstrate the influence of the adapter and LoRA. ``Short'', ``Long'' and ``Comp'' denote short prompts, long prompts, and compositional prompts respectively. The best results are in \textbf{bold}.}
  \label{table: 3}
  \centering
  \scalebox{1}{
  \begin{tabular}{l|cc|ccc}
    \toprule
    & SD & CLIP+U-Net & w/o Adapter & w/o LoRA & T5+U-Net \\
    \midrule
    Short - FID  & \textbf{20.32} & 23.57 & 23.81 & \textbf{22.35} & 23.11 \\
    Short - Aesthetics & \textbf{5.899} & 5.609 & 5.807 & 5.829 & \textbf{5.881} \\
    Short - CLIP Score & \textbf{0.3132} & 0.3102 & 0.3147 & 0.3107 & \textbf{0.3156} \\
    \midrule
    Long - Aesthetics & \textbf{6.120} & 6.003 & 6.131 & 6.273 & \textbf{6.305} \\
    Long - CLIP Score & \textbf{0.3171} & 0.3120 & 0.3106 & 0.3097 & \textbf{0.3193} \\
    \midrule
    Comp - Color & 0.3570 & \textbf{0.3578} & 0.3550 & 0.2485 & \textbf{0.3889} \\
    Comp - Shape & 0.3563 & \textbf{0.3752} & 0.3044 & 0.2944 & \textbf{0.3552} \\
    Comp - Texture & 0.4028 & \textbf{0.4506} & 0.4001 & 0.3190 & \textbf{0.4524} \\
    Comp - Spatial & 0.1225 & \textbf{0.1296} & \textbf{0.1651} & 0.0956 & 0.1582 \\
    Comp - Non-Spatial & \textbf{0.3104} & 0.3009 & 0.3065 & 0.2998 & \textbf{0.3068} \\
    Comp - Complex & \textbf{0.3042} & 0.2985 & 0.2878 & 0.2687 & \textbf{0.3072} \\
    \bottomrule
  \end{tabular}}
\end{table}

\noindent \textbf{LoRA and Adapter}
Here, we investigate the role of LoRA and adapters in LaVi-Bridge. We use T5-Large as the language model and Stable Diffusion V1.4's U-Net as the vision model. For the LoRA experiments, we kept the language and vision models fixed without introducing LoRA, and only trained the adapter. For the adapter experiments, considering the mismatch in the dimensions of text embeddings from the language model and the input embeddings acceptable by the vision model, we aligned the dimensions between the language and vision models using a single linear layer instead of stacked feedforward layers which include non-linear activation layers. Under this setting, we trained both LoRA and this linear layer.

The visualization results are shown in the bottom three rows of \cref{figure: fig6}. We can observe that both image quality and text alignment are significantly affected when LoRA and adapters are not used. For example, in the case of ``Bull Fit Athlete'' in the seventh column, without LoRA or the adapter, the model cannot understand and integrate these two less related elements, and the image quality is much lower compared to results generated by the original setting. We also found that the results without LoRA are worse than those without the adapter. For instance in the fourth column, there is not even a ferret present in the image in the absence of LoRA. 

The right three columns of \cref{table: 3} present the quantitative evaluation results. We find that our default setting, which utilizes both LoRA and the adapter, achieves the best performance in most cases. Additionally, overall, the absence of LoRA has a significant impact on text alignment, with many text alignment evaluation metrics being much lower compared to the absence of the adapter.

\section{Conclusion}
\label{section: Conclusion}

In this paper, we propose LaVi-Bridge, which works on text-to-image diffusion models. LaVi-Bridge is capable of connecting various language models and generative vision models for text-to-image generation. It is highly versatile and can adapt to different structures. LaVi-Bridge is also flexible, as it achieves integration without modifying the original weights of language and vision models. Instead, it utilizes LoRA and an adapter for fine-tuning. Additionally, under LaVi-Bridge, using superior language or vision models can enhance the text comprehension capability or image quality. These advantages enable LaVi-Bridge to help text-to-image diffusion models leverage the latest advancements in the areas of Natural Language Processing and Computer Vision, to enhance text-to-image generation. We believe that this task holds significant research value and requires further exploration. LaVi-Bridge allows designers, artists, and others to flexibly utilize existing language and vision models to achieve their creative goals. It is of utmost importance to avoid misuse and mitigate potential negative social impacts. In practical deployment, it is crucial to standardize its usage, improve model transparency.

\bibliographystyle{splncs04}
\bibliography{main}

\clearpage

\appendix
\section{Long Prompts}
\label{Section: Long Prompts}

In this section, we provided a comprehensive explanation of the evaluation for long prompts mentioned in the main paper. To conduct this evaluation, we utilized the subset of 30k text-image pairs from COCO2014, which was originally used for evaluating short prompts, and employed the Llama-2 to extend the captions within this subset to a length of 20-70 words. 

\begin{figure}[h]
\centering
  \includegraphics[width=1\linewidth]{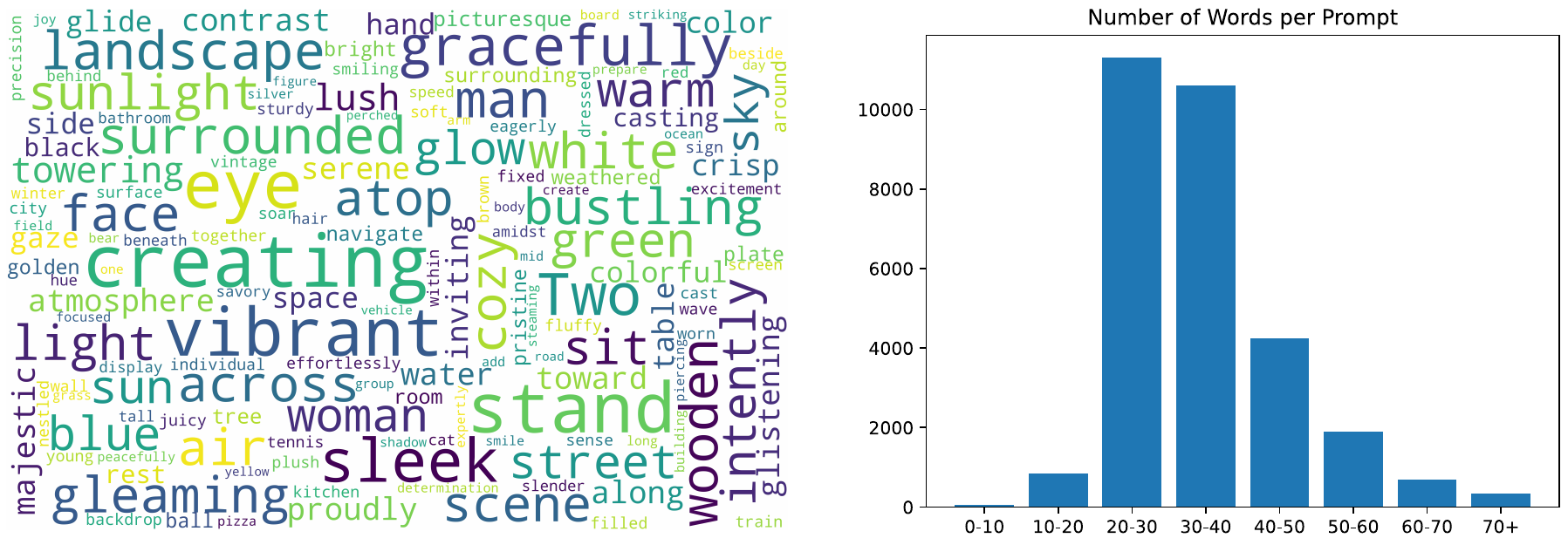}
  \caption{Statistics regarding the long prompts utilized in evaluations in the main paper. The figure on the left visualizes word frequency. The histogram on the right presents the distribution of sentence length, where horizontal axis represents the range of word counts in the prompts, and the vertical axis represents the number of prompts falling within each sentence length range.}
  \label{figure: figs_long_prompts}
\end{figure}

Specifically, we used the Llama-2-7b-chat model and provided the following prompt: \textit{``Please expand the caption to 20-70 words to enrich its semantic meaning: `\textbf{placeholder}'. Just give me one answer.''}  Here, the term ``\textit{\textbf{placeholder}}'' represents the short prompts that we aimed to expand. In this way, we successfully generated 30k long prompts for evaluation. Here are a few examples of the generated long prompts:
\begin{enumerate}
  \item Short Prompt: \textit{``a brown, white and yellow bird standing in the grass.''} $\longrightarrow$ Long Prompt: \textit{``A striking, multi-hued bird with warm brown plumage, crisp white patches on its wings, and vibrant yellow feathers stands gracefully in lush green grass, creating a picturesque scene.''}
  \item Short Prompt: \textit{``two sheep standing in the snow with one looking for food''} $\longrightarrow$ Long Prompt: \textit{``Two fluffy white sheep stand stately in the pristine snow, their wool glistening under the crisp sunlight. One of them eagerly scans the ground, sniffing out potential nourishment amidst the frozen landscape, while the other stands watchful and still, seemingly lost in thought.''}
  \item $\cdots$
\end{enumerate}

We presented statistical analysis of the generated long prompts in the \cref{figure: figs_long_prompts}. The figure on the left visualizes word frequency, while histogram on the right illustrates the distribution of sentence lengths. We observed that the vocabulary diversity in long prompts is quite rich, and the majority of long prompts typically have sentence lengths ranging from 20 to 40 words, and there is also a certain portion that falls within the 40-60 words range.

\section{Training Cost}
\label{Section: Training Cost}

As mentioned in the main paper, LaVi-Bridge does not require modifying the original weights of the language and vision models. Instead, it introduces and trains LoRA and adapters. This approach significantly reduces the training cost compared to training the entire text-to-image diffusion model. For the training of LaVi-Bridge, we utilized 8 A100 GPUs with a batch size of $256$ and completed the training in less than 2 days. Furthermore, we provided a comparison of the number of parameters for different language and vision model combinations in \cref{table: nump}. The leftmost column shows the language and vision models used. It can be observed that training only the LoRA and the adapter leads to a significant reduction in the number of trainable parameters compared to training the original language and vision model. Additionally, both LoRA and the adapter are plug-and-play components, making LaVi-Bridge highly flexible.

\begin{table}[h]
\small
\centering
  \setlength\tabcolsep{2.5pt}
  \caption{Comparison of number of parameters.}
  \label{table: nump}
  \centering
  \scalebox{1}{
  \begin{tabular}{c|ccc|cccc}
    \toprule
    & \multirow{2}{*}[4pt]{Language} & \multirow{2}{*}[4pt]{Vision} & \textbf{\multirow{2}{*}[-1.2pt]{Sum}} & \multirow{2}{*}[-1.2pt]{Adapter} & \multirow{2}{*}[4pt]{Language} & \multirow{2}{*}[4pt]{Vision} & \textbf{\multirow{2}{*}[-1.2pt]{Sum}} \\
    & \multirow{2}{*}[4pt]{Model} & \multirow{2}{*}[4pt]{Model} &  &  & \multirow{2}{*}[4pt]{LoRA} & \multirow{2}{*}[4pt]{LoRA} & \\
    \midrule
    
    \multirow{2}{*}[4pt]{CLIP} & \multirow{2}{*}[-1.2pt]{123M} & \multirow{2}{*}[-1.2pt]{860M} & \textbf{\multirow{2}{*}[-1.2pt]{983M}} & \multirow{2}{*}[-1.2pt]{14M} & \multirow{2}{*}[-1.2pt]{2M} & \multirow{2}{*}[-1.2pt]{28M} & \textbf{\multirow{2}{*}[-1.2pt]{44M}} \\ 
    \multirow{2}{*}[4pt]{U-Net(SD)} &  &  &  &  &  &  & \\ 
    \midrule

    \multirow{2}{*}[4pt]{T5-Small} & \multirow{2}{*}[-1.2pt]{35M} & \multirow{2}{*}[-1.2pt]{860M} & \textbf{\multirow{2}{*}[-1.2pt]{895M}} & \multirow{2}{*}[-1.2pt]{9M} & \multirow{2}{*}[-1.2pt]{0.8M} & \multirow{2}{*}[-1.2pt]{28M} & \textbf{\multirow{2}{*}[-1.2pt]{38M}} \\ 
    \multirow{2}{*}[4pt]{U-Net(SD)} &  &  &  &  &  &  & \\ 
    \midrule

    \multirow{2}{*}[4pt]{T5-Base} & \multirow{2}{*}[-1.2pt]{110M} & \multirow{2}{*}[-1.2pt]{860M} & \textbf{\multirow{2}{*}[-1.2pt]{970M}} & \multirow{2}{*}[-1.2pt]{14M} & \multirow{2}{*}[-1.2pt]{2M} & \multirow{2}{*}[-1.2pt]{28M} & \textbf{\multirow{2}{*}[-1.2pt]{44M}} \\ 
    \multirow{2}{*}[4pt]{U-Net(SD)} &  &  &  &  &  &  & \\ 
    \midrule

    \multirow{2}{*}[4pt]{T5-Large} & \multirow{2}{*}[-1.2pt]{335M} & \multirow{2}{*}[-1.2pt]{860M} & \textbf{\multirow{2}{*}[-1.2pt]{1195M}} & \multirow{2}{*}[-1.2pt]{21M} & \multirow{2}{*}[-1.2pt]{6M} & \multirow{2}{*}[-1.2pt]{28M} & \textbf{\multirow{2}{*}[-1.2pt]{55M}} \\ 
    \multirow{2}{*}[4pt]{U-Net(SD)} &  &  &  &  &  &  & \\ 
    \midrule

    \multirow{2}{*}[4pt]{Llama-2} & \multirow{2}{*}[-1.2pt]{6738M} & \multirow{2}{*}[-1.2pt]{860M} & \textbf{\multirow{2}{*}[-1.2pt]{7598M}} & \multirow{2}{*}[-1.2pt]{229M} & \multirow{2}{*}[-1.2pt]{34M} & \multirow{2}{*}[-1.2pt]{28M} & \textbf{\multirow{2}{*}[-1.2pt]{291M}} \\ 
    \multirow{2}{*}[4pt]{U-Net(SD)} &  &  &  &  &  &  & \\ 
    \midrule

    \multirow{2}{*}[4pt]{T5-Large} & \multirow{2}{*}[-1.2pt]{335M} & \multirow{2}{*}[-1.2pt]{872M} & \textbf{\multirow{2}{*}[-1.2pt]{1207M}} & \multirow{2}{*}[-1.2pt]{30M} & \multirow{2}{*}[-1.2pt]{6M} & \multirow{2}{*}[-1.2pt]{29M} & \textbf{\multirow{2}{*}[-1.2pt]{65M}} \\ 
    \multirow{2}{*}[4pt]{U-Net(LDM)} &  &  &  &  &  &  & \\ 
    \midrule

    \multirow{3}{*}[10pt]{T5-Large} & \multirow{2}{*}[-7pt]{335M} & \multirow{2}{*}[-7pt]{611M} & \textbf{\multirow{2}{*}[-7pt]{946M}} & \multirow{2}{*}[-7pt]{113M} & \multirow{2}{*}[-7pt]{6M} & \multirow{2}{*}[-7pt]{17M} & \textbf{\multirow{2}{*}[-7pt]{136M}} \\ 
    \multirow{3}{*}[10pt]{Transformer} &  &  &  &  &  &  & \\ 
    \multirow{3}{*}[10pt]{(PixArt)} &  &  &  &  &  &  & \\ 
    \bottomrule
  \end{tabular}}
\end{table}

\section{Training Steps}
\label{Section: Training Steps}

As mentioned in the main paper, we trained the LaVi-Bridge for 50k steps. In this section, we present the generated images as training progresses. We utilized the T5-Large as the language model and the U-Net from Stable Diffusion V1.4 as the vision model.

\begin{figure}[h]
\centering
  \includegraphics[width=1\linewidth]{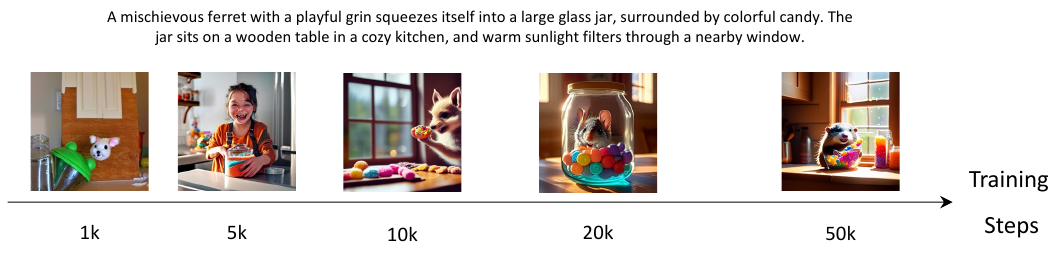}
  \caption{The results for different training steps. Prompts are displayed above, while the number of training steps is shown below.}
  \label{figure: figs_training_step}
\end{figure}

\cref{figure: figs_training_step} illustrates the evolution of the model's performance. Initially, during the first 1k steps, the image quality was poor, and the model struggled to comprehend the given prompt. However, as the training progressed to 10k steps, there was a significant improvement in image quality. By the time it reached 20k steps, the model exhibited enhanced semantic understanding. Finally, at 50k steps, the model demonstrated further optimization compared to the model at 20k step, showcasing the best performance.

\begin{figure}[htb]
\centering
  \includegraphics[width=1\linewidth]{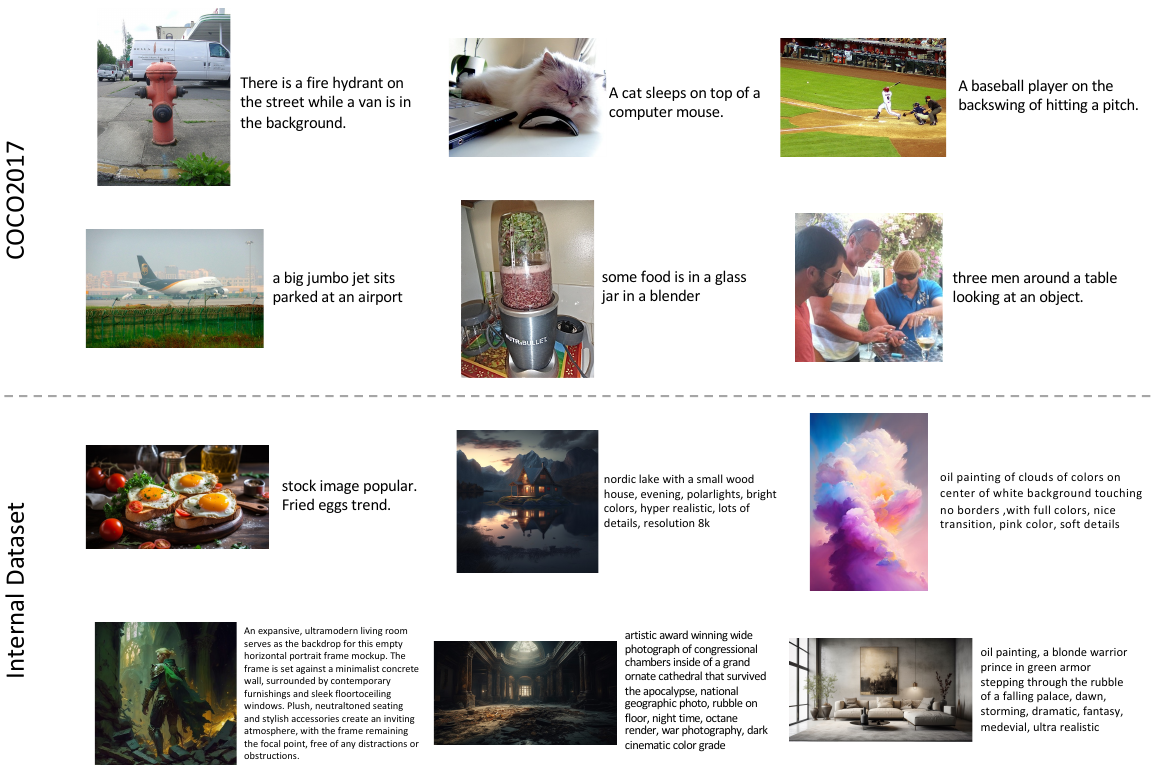}
  \caption{Examples from COCO2017 train set and the internal dataset.}
  \label{figure: figs_trainset}
\end{figure}

\section{Training Set}
\label{Section: Training set}

As mentioned in our main paper, we conducted training using the COCO2017 \cite{lin2014microsoft} train set, which consists of around 600k text-image pairs, along with an additional 400k internal data. The COCO2017 dataset primarily comprises highly realistic images with short and straightforward captions. In order to enhance the diversity and quality of the training data, we collected an additional 400k text-image pairs that exhibit a wide range of artistic styles and high-quality visuals, accompanied by accurate and detailed captions. We provide some examples of the COCO2017 dataset and the internal dataset in \cref{figure: figs_trainset}.

\begin{figure}[h]
\centering
  \includegraphics[width=1\linewidth]{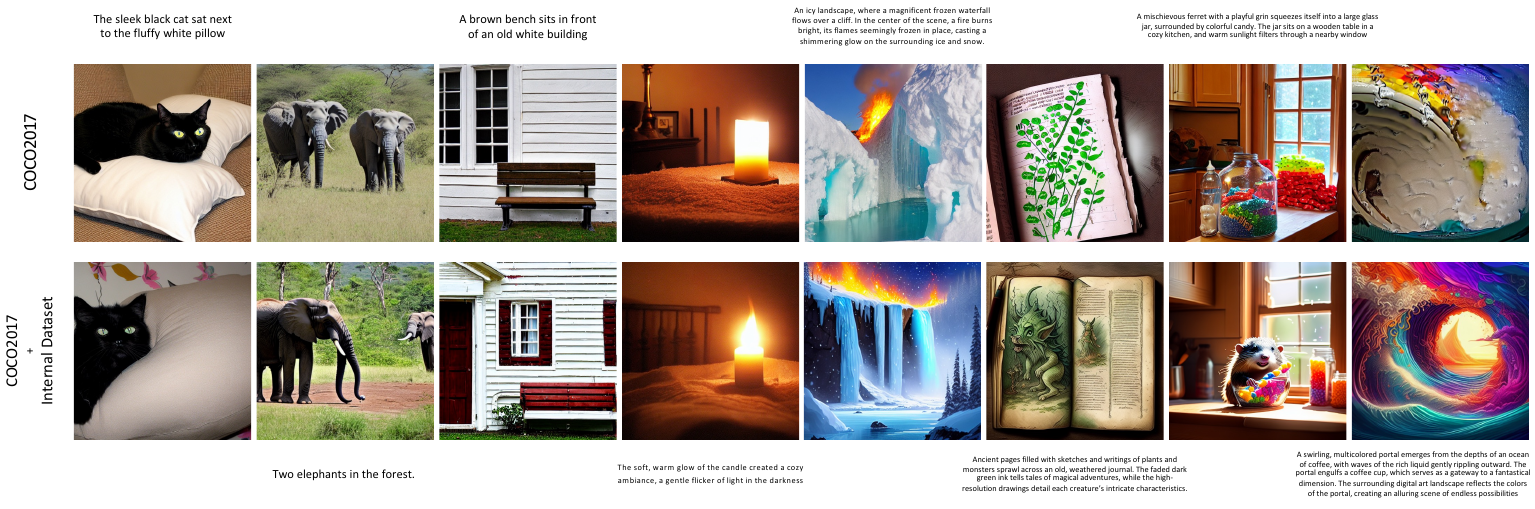}
  \caption{Comparison of training only on COCO2017 and training on both COCO2017 and the internal dataset.}
  \label{figure: figs_comparison}
\end{figure}

In \cref{figure: figs_comparison}, we present a comparison of the results obtained from training on COCO2017 alone and training on both COCO2017 and our internal dataset. We used T5-Large \cite{raffel2020exploring} as the language model and Stable Diffusion V1.4's U-Net \cite{rombach2022high} as the vision model for LaVi-Bridge. It can be observed in \cref{figure: figs_comparison} that in the first three columns, the model trained solely on COCO2017 performs well in terms of both image quality and text alignment. The model accurately understands quantities, as seen in the case of two elephants, and attributes, such as the black cat and white pillow or the brown bench in front of the white building, and so on. Furthermore, the model is capable of generating images using complex prompts, as demonstrated in columns 4-6. However, when the model is tasked with generating images depicting fanciful scenarios, as shown in the seventh column, or images with a non-realistic style, as shown in the last column, the model trained only on COCO2017 struggles to produce such images.

As mentioned earlier, the text-image pairs in COCO2017, from a certain perspective, lack diversity. This is because the images in COCO2017 have a highly realistic style, resulting in a lack of variety, and their quality is relatively low. Additionally, all the captions in COCO2017 are short and very direct, as illustrated in \cref{figure: figs_trainset}. Consequently, it is expected that models trained on COCO2017 will face challenges in generating images of whimsical scenarios or non-realistic styles. Fortunately, there are now many high-quality text-image datasets available, such as \cite{sun2024journeydb}, and we highly recommend incorporating these datasets into training in order to achieve better image generation.

Here, we want to emphasize the contribution of LaVi-Bridge again. LaVi-Bridge is a framework designed to bridge various language and vision models. Even though it was trained only on COCO2017, as can be seen from the first six columns in \cref{figure: figs_comparison}, LaVi-Bridge is still effective. The reason for the poor performance in the last two columns in the first row of \cref{figure: figs_comparison} is due to the limited diversity of the COCO2017 training set as previously mentioned. If LaVi-Bridge is trained on a more general text-image dataset, it will become more versatile, enabling better generation on a wide variety of images.

\section{More Visualization Results}
\label{Section: More Visualization Results}

\begin{figure}[h]
\centering
  \includegraphics[width=1\linewidth]{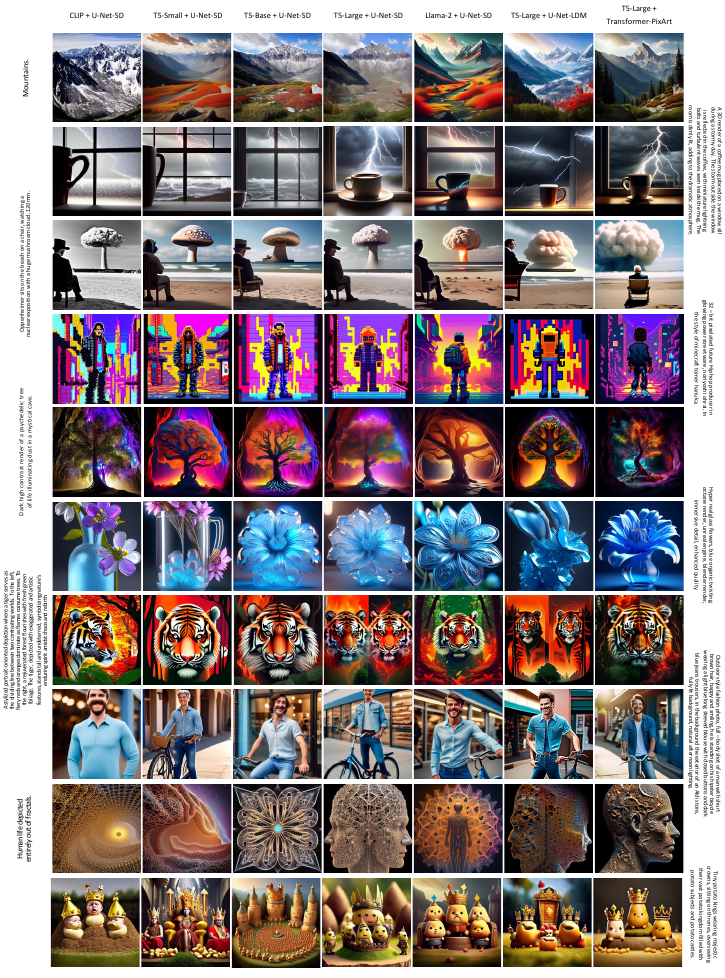}
  \caption{More visualization results. The first column to the seventh column present the results of different combinations using LaVi-Bridge, where the language and vision models used are indicated at the top of each column. The prompts for each row are displayed either on the right or left.}
  \label{figure: figs_more_vis_res}
\end{figure}

In this section, we present additional visualization results in \cref{figure: figs_more_vis_res}. In columns one through five, we kept the vision model fixed as U-Net from Stable Diffusion V1.4 and employed different language models. In columns four, six, and seven, we kept the language model fixed as T5-Large and utilized different vision models.

\end{document}